# Automated Vulnerability Detection in Source Code Using Quantum Natural Language Processing*

Mst Shapna Akter[1], Hossain Shahriar[2], and Zakirul Alam Bhuiya[3]

[1] Department of Computer Science, Kennesaw State University, Kennesaw
makter2@students.kennesaw.edu

[2] Department Information Technology, Kennesaw State University, USA
hshahria@kennesaw.edu

[3] Department of Computer and Information Sciences, Fordham University, New York, USA
zakirulalam@gmail.com

Abstract. One of the most important challenges in the field of software code audit is the presence of vulnerabilities in software source code. Ev-ery year, more and more software flaws are found, either internally in proprietary code or revealed publicly. These flaws are highly likely ex-ploited and lead to system compromise, data leakage, or denial of ser-vice. C and C++ open-source code are now available in order to create a large-scale, classical machine-learning and quantum machine-learning system for function-level vulnerability identification. We assembled a siz-able dataset of millions of open-source functions that point to poten-tial exploits. We created an efficient and scalable vulnerability detection method based on a deep neural network model– Long Short-Term Mem-ory (LSTM), and quantum machine learning model– Long Short-Term Memory (QLSTM), that can learn features extracted from the source codes. The source code is first converted into a minimal intermediate representation to remove the pointless components and shorten the de-pendency. Previous studies lack analyzing features of the source code that causes models to recognize flaws in real-life examples. Therefore, We keep the semantic and syntactic information using state-of-the-art word embedding algorithms such as Glove and fastText. The embedded vectors are subsequently fed into the classical and quantum convolutional neural networks to classify the possible vulnerabilities. To measure the performance, we used evaluation metrics such as F1 score, precision, re-call, accuracy, and total execution time. We made a comparison between the results derived from the classical LSTM and quantum LSTM using basic feature representation as well as semantic and syntactic represen-tation. We found that the QLSTM with semantic and syntactic features detects significantly accurate vulnerability and runs faster than its clas-sical counterpart.

---

* Supported by organization x.

## 1 Introduction

Security in the digital sphere is a matter of increasing relevance. However, there is a significant invasion threat to cyberspace. Security vulnerabilities caused by buried bugs in software may make it possible for attackers to penetrate systems and apps. Each year, thousands of these vulnerabilities are found internally in proprietary code [1].

Recent well-publicized exploits have demonstrated that these security flaws can have catastrophic impacts on society and the economy in our healthcare, energy, defense, and other critical infrastructure systems [2]. For instance, the ransomware Wannacry swept the globe by using a flaw in the Windows server message block protocol [3, 4]. According to the Microsoft Security Response Center, half of 2015 had an industry-wide surge in high-severity vulnerabilities of 41.7%. Furthermore, according to a Frost and Sullivan ( a global growth consulting company) analysis released in 2018, there was an increase in severe and high severity vulnerabilities in Google Project, going from 693 in 2016 to 929 in 2017, with zero coming in second place in terms of disclosing such flaws. On August 14, 2019, Intel issued a warning on a high-severity vulnerability in the software it uses to identify the specifications of Intel processors in Windows PCs

[5] . The paper claims that these defects, including information leaking and de-nial of service assaults, might substantially affect software systems. Although the company issued an update to remedy the problems, an attacker can still use these vulnerabilities to escalate their privileges on a machine that has already been compromised.

A good technique to lessen the loss is early vulnerability discovery. The pro-liferation of open-source software and code reuse makes these vulnerabilities sus-ceptible to rapid propagation. Source code analysis tools are already available; however, they often only identify a small subset of potential problems based on pre-established rules.

Software vulnerabilities can be found using a technique called vulnerability detection. Static and dynamic techniques are used in conventional vulnerability detection [6]. Static approaches evaluate source code or executable code without launching any programs, such as data flow analysis, symbol execution [7], and theorem proving [8]. Static approaches can be used early in software develop-ment and have excellent coverage rates. It has a significant false positive rate. By executing the program, dynamic approaches like fuzzy testing [9], and dy-namic symbol execution [10] can confirm or ascertain the nature of the software. Dynamic methods depend on the coverage of test cases, which results in a low recall despite their low false positive rate and ease of implementation.

The advancement of machine learning technology incorporates new approaches to address the limitations of conventional approaches. One key research direction is developing intelligent source code-based vulnerability detection systems. It can be divided into three categories: using software engineering metrics, anomaly detection, and weak pattern learning. To train a machine learning model, software engineering measures, including software complexity, developer activity, and code commits were initially investigated. This strategy was motivated by the idea that software becomes more susceptible as it becomes more complicated. However, in those works, accuracy, and recall need to be enhanced.

Numerous research organizations are utilizing the potential of quantum machine learning (QML) to handle massive volumes of data with the aid of quantum random access memory (QRAM) [11]. In general, the phrase "Quantum Machine Learning" refers to a field that combines quantum computing, quantum algorithms, and classical machine learning. In this field, real-world machine learning problems are addressed using algorithms that make use of the effectiveness and concepts of quantum computing [12]. Quantum computers have enormous processing capacity due to the core ideas of quantum machine learning, such as quantum coherence, superposition, and entanglement, which pave the way for the increasingly widespread use of quantum computing in technological disci-plines [13]. In contrast to traditional computing, the fundamental building block of quantum computing, the Qubit, can use both 0 and 1 to pursue several computation routes concurrently [14]. A qubit state is a vector in two-dimensional space, shown by the linear combination of the two basis states ($|0>$, and $|1>$) in a quantum system: $|\psi> = \alpha|0> + \beta|1>$, where $\alpha, \beta \in \mathbb{C}$ are probability amplitudes required to satisfy $|\alpha|^2 + |\beta|^2 = 1$. Such a sequence of basis states is referred to as quantum superposition, while the correlations between two qubits through a quantum phenomenon are termed entanglement.

We have come up with a solution for detecting software vulnerabilities using a deep neural network model– Long Short-Term Memory (LSTM), and a quan-tum machine learning model– Long Short-Term Memory (QLSTM), that learns features extracted from the source codes. We first transformed the samples of source code into the minimum intermediate representations through dependency analysis, program slicing, tokenization, and serialization. Later, we extracted se-mantic features using word embedding algorithms such as GloVe and fastText. After finishing the data preprocessing stage, we fed the input representation to the LSTM and QLSTM. We found that QLSTM provides the highest result in terms of accuracy, precision, recall, and f1 score, which are 95, 97, 98, and 99 percent, respectively.

The contribution of this project is as follows:

1. We extracted semantic and syntactic analysis using FastText and Glove word embeddings. Sequential models have the drawback of not understanding the pattern in the text data; it is very tough to catch a similar pattern when unseen data arrives. However, it can be solved by extracting semantic features that can help the models recognize the pattern based on the context, even if unseen data comes.

2. We developed a Quantum Long-short Term Memory algorithm for dealing with sequential data. We have shown how to develop a quantum machine-learning algorithm for detecting software vulnerability.

3. We made a comparative analysis between the results derived from the classical LSTM model and the quantum LSTM model using the basic features as well as semantic and syntactic features.

4. We used both the accuracy and efficiency measurements.

We organize the rest of the paper as follows: We provided a brief background study on classical machine learning and quantum machine learning in section 2. In section 3, we explain the methods we followed for our experimental research. The results derived from the experiment are demonstrated in Section 4. Finally, section 5 concludes the paper.

## 2 Literature Review

Researchers are interested in the recently developed quantum machine learn-ing strategy for identifying and preventing software and cybersecurity vulnera-bilities [1] to address the shortcomings of conventional machine learning tech-niques. For classifying software security activities like malware, ransomware, and network intrusion detection, various machine learning techniques, including Neural Networks, Naive Bayes, Logistic Regression, Recurrent Neural Networks (RNN), Decision Trees, and Support Vector Machines are successfully used. We have tried to go through the classical machine learning and Quantum machine learning-related papers that have been applied to the software security domain.

Previously, Zeng et al. [15] reviewed software vulnerability analysis and dis-covery using deep learning techniques. They found four game changers who contributed most to the software vulnerability using deep learning techniques. Game-changer works describe concepts like– automatic semantic feature extrac-tion using deep learning models, end-to-end solutions for detecting buffer error vulnerabilities, applying a bidirectional Long Short Term Memory (BiLSTM) model for vulnerability detection, and deep learning-based vulnerability detec-tor for binary code.

Yamaguchi et al. [16] showed an anomaly detection technique for taint-style vulnerabilities. It groups the variables that pass on to functions with sensitive security. Then, the violation is reported as a potential vulnerability by anomaly detection. This strategy works well with taint-style vulnerability but not with all vulnerabilities.

Wang et al. [17] proposed an automatic semantic learning process using deep learning models for defect detection in source code. They used DBN, a generative graphical model capable of learning representation that can reconstruct training data with a high probability. The input they provided are [..., if, foo, for, bar, ...] and [..., foo, for, if, bar, ...], respectively. Compared to the traditional features, their semantic features improve WPDP on average by 14.7% in precision, 11.5% in the recall, and 14.2% in F1 score.

Kim et al. [18] proposed a technique for identifying similarity-based vulnera-bilities. Although, this approach is only effective against vulnerabilities brought on by code cloning.

A comparison study based on the effectiveness of quantum machine learn-ing (QML) and classical machine learning was shown by Christopher Haven-stein et al. [19]. The authors worked on reproducible code and applied ML and QML algorithms. Later, quantum variational SVMs were used instead of conven-tional SVMs since they exhibit greater accuracy. The future potential of quantum multi-class SVM classifiers are highlighted in their conclusion.

Quantum machine learning (QML) research was carried out by Mohammad Masum et al. [20] to identify software supply chain attacks. The researchers used a variety of cutting-edge techniques, such as the Quantum Support Vector Machine (QSVM) and Quantum Neural Network, to examine how to accelerate the performance of quantum computing (QNN). The authors discovered two open-source quantum simulators— IBM Qiskit and TensorFlow quantum for software supply chain threats detection. The study's conclusions indicate that quantum machine learning is superior to classical machine learning in terms of computing time and processing speed.

MJH Faruk et al. [21] studied quantum cybersecurity from both threats and opportunity perspectives. The authors have provided a comprehensive review of state-of-the-art quantum computing-based cybersecurity approaches. The re-search indicated that quantum computing can be utilized to address software security, cybersecurity, and cryptographic-related concerns. On the other hand, the malicious individual also misuses quantum computing against software in-frastructure due to the immense power of quantum computers.

Payares and Martinez-Santos [22] demonstrated the importance of using gen-eralized coherent states for the SVM model. SVM is a classical machine learning model, and coherent states are a calculational tool here. They used the RKHS concept to connect the thread in the SVM model. Such reproducing kernels are responsible for the overlapping situation of canonical and generalized coherent states. Canonical coherent states regenerate the radial kernels while the POVM calculates the overlap functions, which eventually decreases the computational times when the high dimensional feature problem arises. The Quantum version of SVM recently played an important role in solving classification and detection problems, but the quantum version of LSTM is unexplored.

Khan and Pal [23] proposed a quantum computing-based technique for devel-oping reverse engineering gene regulatory networks using a time-series gene ex-pression dataset. The model they used is recurrent neural networks. The method they used is comparatively new, and the results are satisfactory. They applied a 4-gene artificial genetic network model as part of experimental work and proposed a 10-gene and 20-gene genetic network. They found that quantum computing provides very good accuracy and it reduces the execution time.

Choi et al. [24] proposed a brief description of Quantum Graph Recurrent Neural Networks (QGRNN). Previously, several works have been done in the machine learning field with some optimized models. Graph Neural Networks is a subfield of machine learning which came to the focus. Researchers developed a quantum graph neural network using circuits that convert the bits into qubits. Among the other quantum machine learning models, the quantum graph recur-rent neural network (QGRNN) is proven to be more effective. They proposed a Variational Quantum Eigensolver (VQE) for converting the bits into quantum states and the converting them into QGRNN.

Oh et al. [25] provided a survey on Quantum Convolutional Neural Network (QCNN). QCNN is a quantum version of classical CNN. It is sometimes very challenging when data comes with high dimensions and thus makes the model very large. But the quantum version of CNN overcomes such issues and improves performance, and makes the model efficient. The first study proposes a quantum circuit that uses Multi-scale Entanglement Renormalization Ansatz (MERA). The second work they showed used a hybrid learning model, which adds a quantum convolution layer.

Lu et al. [26] proposed a Quantum Convolutional Neural Network to find the optimal number of convolutional layers. They used CIFAR10 for training VGG-19 and 20-layer, 32-layer, 44-layer, and 56-layer CNN networks. Then they compare the difference between the optimal and non-optimal CNN models. They found that the accuracy drops from 90% to 80% as the layers increases to 56 layers. However, the CNN with optimization made it possible as the accuracy is more than 90%, and the parameters are reduced by half. Their experiment indicates that the proposed method can improve the network performance degra-dation very well, which causes by hidden convolutional layers, and reduce the computing resources.

## 3 Methodology

We adopted a Quantum Long Short-Term Memory (QLSTM), a subfield of Quantum Machine Learning (QML), for this research and applied the model to the text dataset. Figure 1 demonstrates the framework representing the implementation process. At first, we pre-processed raw data prior to providing it as input to the QML model. We used Python, keras tokenizer, sklearn LabelEncoder, Keras sequence, Keras padding, Keras embedding, Glove, and FastText embeddings for pre-processing the dataset. Keras library has been used to extract the basic input representation and the semantic and syntactic representations have been extracted using Glove and FastText Embeddings. For the experiment, we consider only the balanced portions of the dataset, which contains an almost equal number of vulnerable and non-vulnerable datasets, to avoid underfitting or overfitting. In quantum machine learning models, we need to feed numerical values, so we converted text data into numerical values, and all the numerical values were normalized to maintain a similar scale. We made a comparison with results derived from LSTM and QLSTM models with or without training the semantic and syntactic representations. Results have been shown in section 4.

The sigmoid layer state's equation is:

### 3.1 Dataset Specification

From the standpoint of the source code, the majority of flaws originate in a crucial process that poses security risks. The function, assignment, or control statement considers the primary operation. An adversary can affect this crucial operation either directly or indirectly by manipulating certain factors or cir-cumstances. A large number of training examples are required to train machine learning models that can effectively learn the patterns of security vulnerabilities directly from code. We chose to analyze software packages at the function-level because it is the lowest granularity level that captures a subroutine's overall flow. We compiled a vast dataset of millions of function level examples of C and C++ code from the SATE IV Juliet Test Suite, Debian Linux distribution, and public Git repositories on GitHub described in the paper of Russel [27]. In this project we have used CWE-119 vulnerability feature. CWE-119 indicates issues associated with buffer overflow vulnerability. A buffer overflow occurs when data is written to the buffer that is longer than the buffer itself, overwriting storage units outside the buffer in the process. According to a 2019 report from Com-mon Weakness Enumeration, the buffer overflow vulnerability has evolved into the vulnerability that has been most adversely affected. Although we have used the buffer overflow issue as an example, our method can be used to find other kinds of vulnerabilities as well. Figure 1 illustrates an intra-procedural buffer overflow vulnerability.

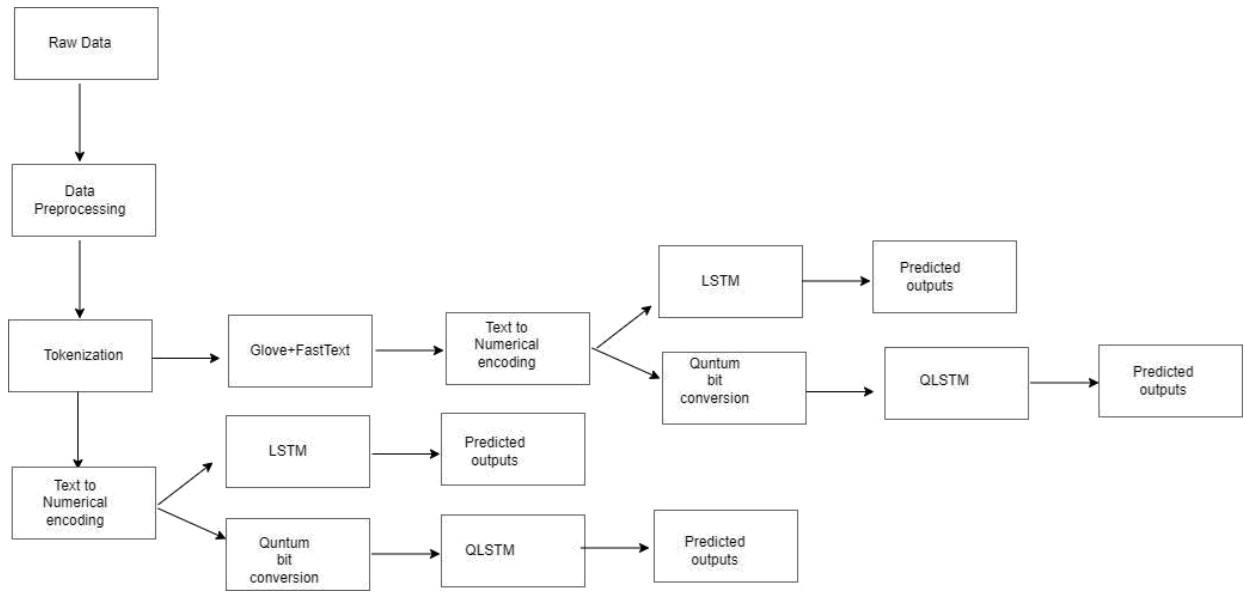

Fig. 1: Framework representation of the implementation process.

The data we have used in this project can be found in this link: https://cwe.mitre.org/data/definitions/119.html. The dataset has a subfolder: train, validation, and test. Each of the folders contains a CSV file. The CSV file stores the text data and corresponding labeling. We analyzed the dataset

```
1   void bar(char *buf, char *src) {
2       strcpy(buf, src);
3   }
4   int main() {
5       char buf[10];
6       char src[10];
7       memset(src, 'A', 10);
8       src[10 - 1] = '\0';
9       bar(buf, src);
10      for (int i = 0; i <= 10; i++)
11      //writes buf [10] and overruns memory
12          buf[i] = 'B';
13      return 0;
14  }
```

Fig. 2: An intra-procedural buffer overflow vulnerability.

and found some common words with it's corresponding counts: = (505570), if (151663), n (113301), == (92654), return (77438), * (71897), the (71595), int (53673), < (43703), + (41855), for (35849), char (33334), else (31358).

3.2  Input Representation

The basic input representation technique is representing each word with a num-ber, eventually creating a pattern. The neural networks sometimes fail to learn in-depth patterns and thus not giving accurate results for real-life examples. The biased nature of data occurs in this case, and the model can recognize data only with the same pattern. In real life, all data are not the same. To overcome this issue, Glove and fasTest have been developed to extract the semantic and syn-tactic features. Even with different patterns, a model trained with semantic and syntactic features is able to recognize the vulnerability with the help of context of words.

Tokenizer In the natural language processing project, the basic units called words are mandatory to concentrate on the computational process. In the NLP field, the word is also known as a token. Tokenization is a process that sepa-rates sentences into small units that can be used for several purposes. With this concept, Keras provides a tokenizer library called Tokenizer. Tokenizer contains two methods named tokenize() and detokenize(). The methods go through the plain text and separate the word. We used the Keras tokenizer for our initial data preprocessing step kathuria2019real.

GloVe Semantic vector space language algorithms replace each word with a vector. The vectors are useful since the machine cannot understand words but a vector. Therefore, numerous applications can make use of these vectors as

features– question answering [28], parsing [29], document classification [30], in-formation retrieval, and named entity recognition [31]. The glove is a language algorithm for prevailing vector representations of words. This is an unsupervised learning algorithm; the training process has been performed on global word-word co-occurrence statistics from a corpus [32].

Pennington et al. [32] dispute that the online scanning process followed by word2vec is inferior as it does not provides global statistical values about word co-occurrences. The model produces a vector space with a valid substructure with 75% of performance on a real-life example. GloVe was built on two concepts– lo-cal context window and global matrix factorization. CBOW and skip-Gram are Local context window methods. CBOW is better for frequent words, whereas Skip-gram works well on small datasets with rare observations. While global matrix factorization is the matrix factorization method that derives from lin-ear algebra is responsible for reducing the long-term frequency matrices. The matrices constitute the occurrence of words.

fastText FasText is an embedding method that uses a word's deep-down structure to improve the vector representations acquired from the skip-gram method. Modifications such as sub-word generation and skip-gram with negative sampling happen in the skip-gram method to develop the fasText model.

Sub-word generation: For a specific word, it generates character n-grams of length 3-6. The source code mostly uses words such as return, char, else, and int shown in Table 1 and Figure 3. The first step would be to take a word and add an angular bracket. It denotes the beginning and end of the word. For instance Return , char , int . This approach conserves the interpretation of short words that may come up as n-grams of other words. Moreover, it captures the meaning of suffixes and prefixes. The length of n-grams can be controlled by using -maxn and -minn flags for the maximum and the minimum number of characters [33]. The fastText model is somewhat considered a bag of words model aside from the n-gram window selection. No internal structure is present for the featurization of the words. However, the n-gram embeddings can be turned off by setting it as 0, which is useful for some particular words which do not bring any meaning in the entire corpus. The purpose of putting ids in words is while updating the model, fastText learns weights for each of the n-grams as well as the entire word token [34].

### 3.3 Classification Models

The vector representation of the software source code has been fed to the LSTM, and QLSTM models. The dataset has been divided into training and validation portions for the purpose of training the models. Finally, the test dataset is used to evaluate each trained model.

Long Short-Term Memory (LSTM) LSTM is A popular artificial Recur-rent Neural Network (RNN) model, which works with datasets that preserve sequences such as text, time series, video, and speech. Using this model for se-quential datasets is effectiveSince as it can handle single data points. It follows the Simple RNN model's design and an extended version of that model. How-ever, unlike Simple RNN, it has the capability of memorizing prior value points since it developed for recognizing long-term dependencies in the text dataset. Three layers make up an RNN architecture: input layer, hidden layer, and output layer mandic2001recurrent, akter2022forecasting. Figure 3 depicts the LSTM's structural layout. The elementary state of RNN architecture is shown as the mathematical function:

$$h_t = f(h_{t-1}, x_t; \vartheta) \tag{1}$$

Here, $\vartheta$ denotes the function parameter, $h_t$ denotes the current hidden state, $x_t$ denotes the current input, $f$ denotes the function of previous hidden state.

When compared to a very large dataset, the RNN architecture's weakness is the tendency to forget data items that are either necessary or unneeded. Due to the nature of time-series data, there is a long-term dependency between the current data and the preceding data. The LSTM model has been specifically developed to address this kind of difficulty. It is first proposed by Hochreiter Long [35].

This model's main contribution is its ability to retain long-term dependency data by erasing redundant data and remembering crucial data at each update step of gradient descent [36]. The LSTM architecture contains four parts: a cell, an input gate, an output gate, and a forget cell [37].

The purpose of the forget cell is to eliminate extraneous data by determining which data should be eliminated based on the state (t) − 1 and input x(t) at the state c(t) − 1.

At each cell state, the sigmoid function of the forget gate retains all 1s that are deemed necessary values and eliminates all 0s that are deemed superfluous.[35]. The forget gate state's equation is stated as follows:

$$f_t = \sigma(W_f \cdot [h_{t-1}, x_t] + b_f) \tag{2}$$

where $f_t$ denotes sigmoid activation function, $h_{t-1}$ denotes output from previ-ous hidden state, $W_f$ denotes weights of forget gate, $b_f$ denotes bias of forgetting gate function, and finally $x_t$ dentoes current input.

After erasing the unneeded value, new values are updated in the cell state. Three steps make up the procedure: The first step is deciding which values need to update using sigmoid layer called the "input gate layer". Second, creating a vector of new candidate values using the tanh layer. Finally, steps 1 and 2 are combined together to create and update the state.

The equation for the sigmoid layer is as follows:

$$i_t = \sigma(W_i \cdot [h_{t-1}, x_t] + b_i) \tag{3}$$

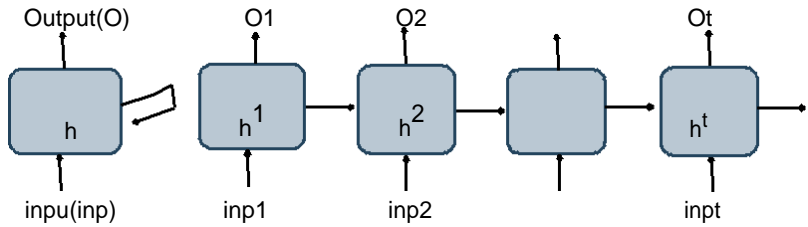

Fig. 3: LSTM neural network structure.

The tanh layer generates a vector of new candidates for producing a new value to the state $C(t)$, while the sigmoid layer determines which value should be updated.

The equation for tanh layer's equation is as follows:

$$\tilde{C}(t) = \tanh(W_C \cdot [h_{t-1}, x_t] + b_C) \tag{4}$$

The addition of $\tilde{C}(t) * i_t$ and $C_{t-1} * f_t$ updates the new cell at state $C(t)$. The updated state's equation is:

$$C_t = C_{t-1} * f_t + \tilde{C}(t) * i_t \tag{5}$$

In order to determine which output needs to be maintained, the output is ultimately filtered out using the sigmoid and the tanh functions.

$$O_t = \sigma(W_O \cdot [h_{t-1}, x_t] + b_O) \tag{6}$$

$$h_t = O_t * \tanh(C_t) \tag{7}$$

In this state, $h_t$ gives outputs that are used for the input of the next hidden layer.

Quantum Long Short-Term Memory (QLSTM) We propose a novel frame-work of a modified version of the recurrent neural network- Long short-term memory (LSTM) with Variational Quantum Circuits (VQCs), basic functions discussed briefly in the following section. The model has been implemented us-ing a hybrid quantum-classical approach, which is fitted for NISQ devices as the approach utilizes the greater expressive power during the iterative optimization process provided by the quantum entanglement.

Variational Quantum Circuits (VQCs): A quantum circuit with tun-able parameters. We choose the circuit to use in the NISQ device as it is robust against quantum noise [38]. Moreover, VQCs are more intensive than classical neural networks as it has the ability to represent certain functions with a lim-ited number of parameters [39]. Recurrent Neural Networks approximate any computable function, even with one single hidden layer; Therefore, pairing the

LSTM model with the VCQs enables the learning process very fast [40]. It has been successfully experimented on several tasks such as classification [41], func-tion approximation [42], generative modeling [43], deep reinforcement learning [44], and transfer learning [45]. The architecture for this circuit is demonstrated in Figure 4.

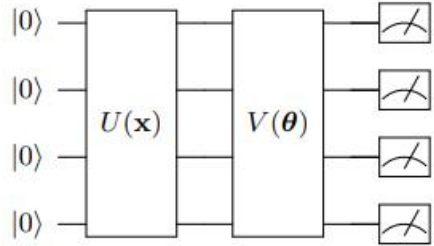

Fig. 4: Variational Quantum Circuit's architecture

The U(x) block represents the state preparation which converts the classical input x into the quantum state. In contrast, the block represents the variational part along with the learnable parameters for doing optimization during the gra-dient descent process. We measure a subset of the encoded qubits to retrieve a classical bit string, for example, 0100.

Quantum LSTM We modify the traditional LSTM architecture into a quantum version by replacing the neural networks in the LSTM cells with Vari-ational Quantum Circuits (VQCs) [40]. The VQCs play roles in both feature extraction as well as data compression. The components we used for the VCQs are shown in Figure 5.

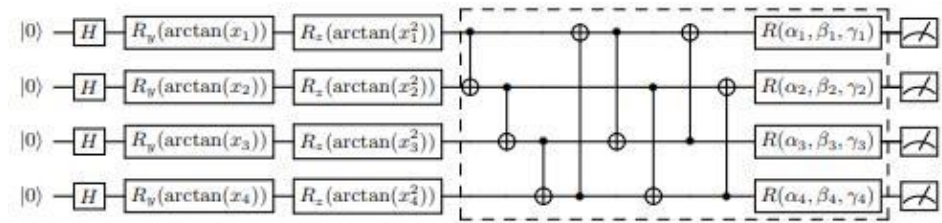

Fig. 5: Components for VQCs

The circuit consists of three parts: data encoding, variational layer, and quan-tum measurements. Data encoding transforms the classical data into a quantum state. The variational layer updates itself during gradient descent and plays the optimization role [40]. The quantum measurements retrieve the values for fur-ther processing. The expected values have been calculated using the simulation software, as we do not have access to the real quantum computer. We have shown the Quantum LSTM architecture in Figure 6.

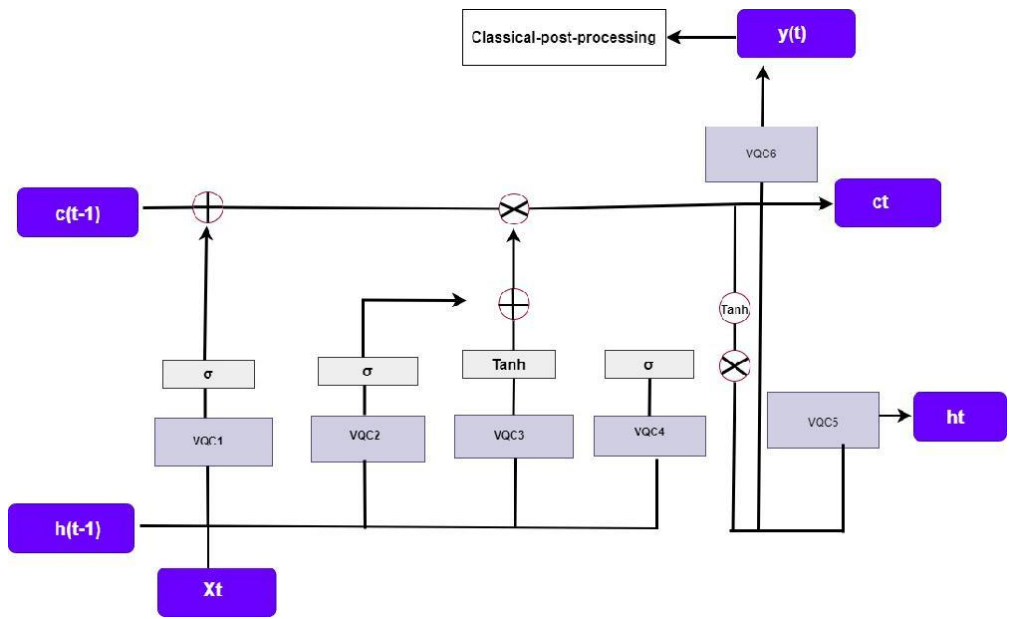

Fig. 6: QLSTM Architecture

$x(t)$ refers to the input at time $t$, $h_t$ refers to the the hidden state, $c_t$ refers to the cell state, and $y_t$ refers to the output. The blocks $\sigma$ and tanh represent the sigmoid and the hyperbolic tangent activation function, respectively. Finally, the $\otimes$ and $\oplus$ represents element-wise multiplication and addition. The mathematical functions [40] for each state of QLSTM model are stated below :

$i_t = \sigma(VQC2(v_t))$
$C_t' = Tanh(VQC3(v_t))$
$c_t = f_t * c_{t-1} + i_t * C_t'$
$o_t = \sigma(VQC4(v_t))$
$h_t = \sigma(VQC5(o_t * tanh(c_t))$
$y_t = VQC6(o_t * tanh(c_t))$

### 3.4 Evaluation metrics

Evaluating a model's performance is necessary since it shows how close the model's predicted outputs are to the corresponding expected outputs. The eval-uation metrics are used to evaluate a model's performance. However, the evalu-ation metrics differ with the types of models. The types of models are classifi-cation and regression. Regression refers to the problem that involves predicting a numeric value. Classification refers to the problem that involves predicting a discrete value. The regression problem uses the error metric for evaluating the models. Unlike the regression problem model, the classification problem uses the accuracy metric for evaluation. Since our motive is to detect the vulnerabili-ties, we used accuracy, f1 score, Precision, and Recall for our evaluation metric depto2021automatic .

Precision : When the model predicts positive, it should be specified that how much the positive values are correct. Precision is used when the False Positives are high. For vulnerability classification, if the model gives low Precision, then many non-vulnerable codes will be detected as flaws; for high Precision, it will ignore the False positive values by learning with false alarms. The Precision can be calculated as follows:

$$Precision = \frac{TP}{TP+FP} \qquad (8)$$

Here TP refers to True Positive values and FP refers to False Positive values.
Recall : The metric recall is the opposite of Precision. The Precision is used when the false negatives (FN) are high. In the vulnerability detection classification problem, if the model gives low recall, then many vulnerable codes will be said as non-vulnerable; for high recall, it will ignore the false negative values by learning with false alarms. The recall can be calculated as follows:

$$Recall = \frac{TP}{TP+FP} \qquad (9)$$

F1 score: F1 score combines Precision and recall and provides an overall accu-racy measurement of the model. The value of the F1 score lies between 1 and 0. If the predicted value matches with the expected value, then the f1 score gives 1, and if none of the values matches with the expected value, it gives 0. The F1 score can be calculated as follows:

$$F1score = \frac{2 \cdot precision \cdot recall}{precision + recall} \qquad (10)$$

Accuracy : Accuracy determines how close the predicted output is to the actual value.

$$Accuracy = \frac{TP+TN}{TP+TN+FP+FN} \qquad (11)$$

here, TN refers to True Negative and FN refers to False Negative.

## 4 Result and Discussion

From the previous study, we found that the application of quantum models has not been applied to the software security field. Since, the majority of software companies face a surge due to software flaws, those require a system that can provide an accurate result as well as efficiency. Before training the neural networks, several criteria need to be followed. One important criterion is feature analysis. There is a huge chance that a classifier performs poorly due to the lack of feature analysis techniques. As investigators did not consider the in-depth feature analysis process in software security field, we have shown a step-by-step process for extracting the semantic and syntactic features. We developed the LSTM model with the same number of parameters for both the classical and quantum versions to get a clear observation. We implemented the classical LSTM architecture using TensorFlow with 50 hidden units.

It has a softmax layer to convert the output to a single target value $y_t$. The total number of parameters is 123301 in the classical LSTM. In case of QLSTM, we used 6 VQCs shown in Figure 5. Each of the VQCs consists of 4 qubits with 2 depths in each variational layer. Additionally, there are 2 parameters for scaling in the final step. The total number of parameter is 122876. We chose pennylane as our simulation environment for the quantum circuit. Through our experimental results, we found that the QLSTM learns faster than the classical LSTM does with a similar number of parameters. Our comparative analysis between the classical Long Short-Term Memory model and the quantum long Short-Term Memory model illustrates in Table 1 and Table 2.

Table 1: Vulnerable Source code Classification results using classical LSTM and Quantum LSTM with no word embedding algorithms

| Models   | Accuracy | precision | Recall | F1 Score | Execution Time |
|----------|----------|-----------|--------|----------|----------------|
| 1* LSTM  | 0.90     | 0.90      | 0.90   | 0.92     | 40min 2s       |
| 1* QLSTM | 0.94     | 0.99      | 0.98   | 0.99     | 13min 52s      |

Table 2: Vulnerable Source code Classification results using classical LSTM and Quantum LSTM with embedding algorithms GloVe + fastText

| Models   | Accuracy | precision | Recall | F1 Score | Execution time |
|----------|----------|-----------|--------|----------|----------------|
| 1* LSTM  | 0.92     | 0.93      | 0.95   | 0.97     | 42min 13s      |
| 1* QLSTM | 0.95     | 0.97      | 0.98   | 0.99     | 19min 11s      |

Table 1 shows the result derived from LSTM and QLSTM with basic input representation, whereas Table 2 shows the result derived from LSTM and QLSTM with semantic and syntactic representation. For both cases, we found that the Quantum version of the LSTM model provides better accuracy and comparatively runs faster. LSTM with basic representation provides 90 percent accuracy, 90 percent precision, 90 percent recall, and 92 percent F1-Score. In comparison, QLSTM with the same basic input representation provides 94 percent Accuracy, 99 percent precision, 98 percent recall, and 99 percent of the F1 score. Classical LSTM with semantic representation provides 92 percent accuracy, 93 percent precision, 95 percent Recall, and 97 percent F1 score. While quantum LSTM with semantic representation provides 95 percent accuracy, 97 percent precision, 98 percent recall, and 99 percent f1 score. We also found that by using the Glove

+ FastText word embedding model, the accuracy improves, but the execution time becomes higher. LSTM with basic input representation, LSTM with se-mantic and syntactic representation, QLSTM with basic input representation, QLSTM with semantic and syntactic representation takes 40min 2s, 42min 13s, 13min 52s, and 19min 11s, respectively. Further, we explored the learning capa-bility of the periodic function of our QML model. The sine function: $y = \sin(x)$

(3) is easier to represent and more effective for comparing. The result is shown in Figure 7. We used 30 epochs, and the models iterated through each model and learned. In each epoch, it learns and optimizes in different batches of inputs. Therefore we can make an observation by comparing the learning representation of the first and last epochs. The vertical blue line represents actual data, and the red line represents the predicted output.

From the sinusoidal graph, we found that QLSTM learns better than LSTM. We also point out that QLSTM learns significantly more information from the first epoch while LSTM gradually learns and improves in the last epoch, still lacking the proper learning capability.

# 1 Conclusion

Quantum computing has recently gained prominence with prospects in the computation of machine learning algorithms that have addressed challenging problems. This paper conducted a comparative study on quantum Long Short-Term Memory (QLSTM) and traditional Long Short-Term Memory (LSTM) and an-alyzes the performance of both models using vulnerable souce code . Moreover, We extracted the semantic and syntactic information using state-of-the-art word embedding algorithms such as Glove and fasText, which can make more accurate result. The QML model was used on a open sourced Penny Lane simulator due to the limited availability of the quantum computer. We have tried to implement machine learning algorithms for sequence modeling, such as natural language processing, vulnerable source code recognition on noisy intermediate-scale quantum (NISQ) device. We assessed the model's performance using accuracy and processing criteria. According to the experimental findings, the QLSTM with Glove and fastText embedding model learns noticeably more vulnerable source

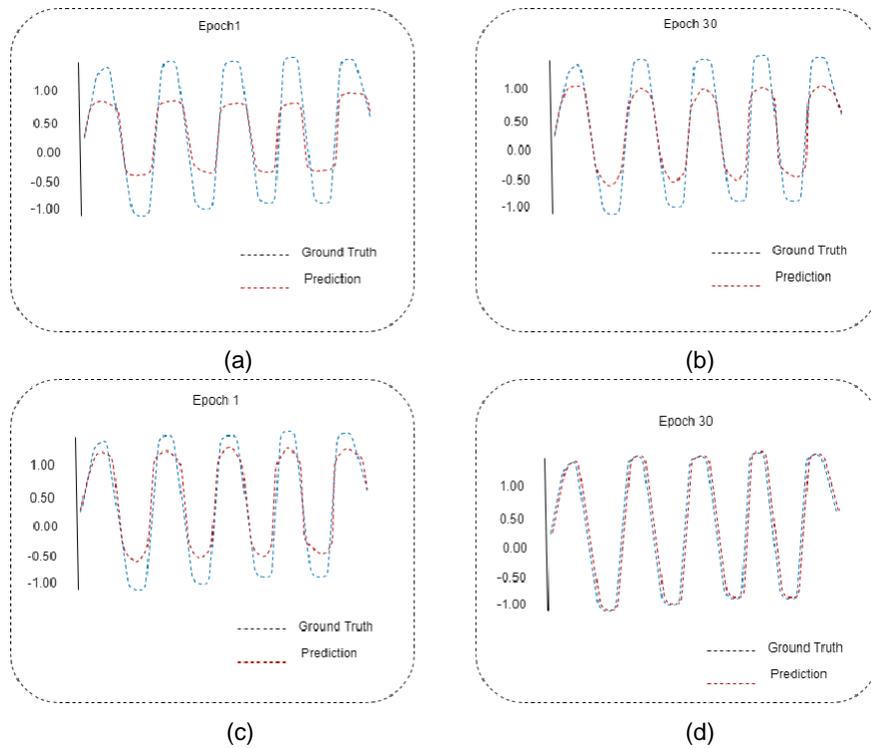

Fig.7: Comparison between sine function learning result obtained from jupyter notebook using LSTM+ Glove + FasText model for (a) epoch 1 and b) epoch 30 and QLSTM+ Glove + FasText model for (c) epoch 1 and (d) epoch 30

code features and operates more quickly than its conventional equivalent. Although advances have been made in quantum machine learning over the past few decades, more work is still needed because the current generation of quan-tum simulators only has a small number of qubits, making them unsuitable for sensitive source code. It is possible that a large number of convergent qubits using quantum machine learning models will have a significant impact on clas-sification performance and computing time.

## Acknowledgement

The work is partially supported by the U.S. National Science Foundation Award #2100115. Any opinions, findings, and conclusions or recommendations expressed in this material are those of the authors and do not necessarily reflect the views of the National Science Foundation.